\documentclass[runningheads, envcountsame, a4paper]{llncs}
 \usepackage{graphicx}
\usepackage{epstopdf}
\usepackage[misc]{ifsym}
\usepackage{graphicx,makecell,bm}
\usepackage[ruled,linesnumbered]{algorithm2e}
\usepackage[hyphens]{url}
\begin{document}
	\title{PathSAGE: Spatial Graph Attention Neural Networks With Random Path Sampling}
	%
	%
	\author{Junhua Ma\inst{1} \and
		Jiajun Li\inst{2} \and
		Xueming Li\inst{1}\Letter \and
		Xu Li\inst{1}}
	%
	%
	\institute{College of Computer Science,	Chongqing University,\\
		  Chongqing 400044, China\\
		\email{$\{$majunhua, lixuemin$\}$@cqu.edu.cn, leuio@foxmail.com}\and
	School of Information
	Technology \& Electrical Engineering, \\
		The University of Queensland, St Lucia Qld 4072, Australia \\
		\email{jiajun.li1@uqconnect.edu.au}}
	\toctitle{PathSAGE: Spatial Graph Attention Neural Networks With Random Path Sampling}
	\tocauthor{Junhua~Ma, Jiajun~Li, Xueming~Li and Xu~Li}
	\titlerunning{PathSAGE}
	\authorrunning{J. Ma, J. Li, X Li and X. Li}
	\maketitle              
	 \setcounter{footnote}{0}
	\begin{abstract}
		Graph Convolutional Networks (GCNs) achieve great success in non-Euclidean structure data processing recently. In existing studies, deeper layers are used in CCNs to extract deeper features of Euclidean structure data. However, for non-Euclidean structure data, too deep GCNs will confront with problems like ''neighbor explosion'' and ''over-smoothing'', it also cannot be applied to large datasets. To address these problems, we propose a model called PathSAGE, which can learn high-order topological information and improve the model's performance by expanding the receptive field. The model randomly samples paths starting from the central node and aggregates them by Transformer encoder. PathSAGE has only one layer of structure to aggregate nodes which avoid those problems above. The results of evaluation shows that our model achieves comparable performance with the state-of-the-art models in inductive learning tasks. 
	\end{abstract}
	\section{Introduction}
	Convolutional Neural Networks (CNNs) have been successfully used in various tasks with Euclidean structure data in recent years. For non-Euclidean structure datasets, graph convolutional networks (GCNs) use the same idea to extract the topological structure information.
	
	\cite{bruna2014spectral} firstly proposed two GCN models with spectral and spatial construction respectively. The spectral model using a Laplacian matrix to aggregate neighborhood information of each node in a graph. The spatial model partition graph into clustering and update them by aggregating function. In order to extract deeper features, model based on CNNs usually deepen the model's layers. While in GCNs, Deepening layers will cause a lot of problems. In spectral construction method, too many layers lead to \textbf{``over smoothing''}\cite{li2018deeper,zhou2018graph}: the features of nodes in graph will tend to be the same. In spatial construction method, it will cause exponential growth of the number of sampled neighbor nodes, called \textbf{``neighbor explosion''}. Node sampling and layer sampling\cite{hamilton2018inductive,chen2018fastgcn,gao2018large,huang2018adaptive,ying2018hierarchical} were proposed to handle this problem, but due to incomplete sampling, the inaccuracy of nodes' representation accumulates errors between layers.
	
	To address these two problems, in this paper, we propose a model called PathSAGE, which can learn high-order topological information by expanding the receptive field.
	Firstly, We design a path sampling technique based on random walk to sample paths starting from central nodes with different lengths, then the sequences of paths are fed into Transformer encoder\cite{vaswani2017attention}, which can extract the semantic and distance information in sequence effectively. As shown in Fig. 1, We view the sequences of paths from the tail nodes to the central node as the central node' neighbors. Secondly, following this idea, we take the average of paths of the same length as the representation of the central node in this level of reception field. Finally, after concatenating the aggregated features of paths with different lengths, the final representation of the central node are used for downstream tasks.
	
	For the two problems mentioned earlier, on the one hand, the aggregation of the central node only perform once in training of a sample, which never cased ``over-smoothing''. On the other hand, all the paths were sampled with a fixed number for representing the central node in our model, instead of recursively sampling exponentially growing neighbor nodes. And each path only contributes to the central node, we do not need to calculate and store the temporary representations of nodes from the middle layer. Furthermore, it prevents the error propagation caused by incomplete sampling. 
	
	Our contribution can be summarized in three points:
	
	\begin{itemize}
		
		\item We utilize the path sampling to take place of the node sampling to avoid error accumulation caused by incomplete node sampling.
		
		\item We propose and evaluate our model Path-SAGE to solve the existing ``neighour explosion'' and ``over-smoothing'' problems. the model can capture richer and more diverse patterns around the central node with only one layer of structure to the central node.
		
		\item We evaluate our model on three inductive learning tasks, and it reaches the state-of-the-art performance on two of them. We analyze the attention weights of Transformer encoder and detect some patterns in the attention mechanism, which can further illustrate how the model works.
		
	\end{itemize}
	\section{Related work}
	GNNs model was initially proposed by \cite{bruna2014spectral}, and the convolution operation in the traditional Euclidean structure was introduced into the graph network with the non-Euclidean structure in this article. They \cite{bruna2014spectral} divided GNN into two construction methods: spectral construction and spatial construction. Subsequently, many studies are carried out around these two aspects. 
	
	In spectral construction, \cite{defferrard2017convolutional} used Chebyshev polynomials with learnable parameters to approximate a smooth filter in the spectral domain, which improves computation efficiency significantly. \cite{kipf2017semisupervised} further reduced the computational cost through local first-order approximation. In spatial construction, MoNet was proposed in \cite{monti2017geometric}, developing the GCN model by defining a different weighted sum of the convolution operation, using the weighted sum of the nodes as the central node feature instead of the average value. \cite{hamilton2018inductive} attempted various aggregator to gather the features of neighbor nodes. \cite{velivckovic2017graph,zhang2018gaan} defined the convolution operation with a self-attention mechanism between the central node and neighbor nodes, \cite{liu2019geniepath} brought the LSTM(Long Short Term Memory networks)\cite{hochreiter1997long} from NLP (Natural Language Processing) to GNN, and built an adaptive depth structure by applying the memory gates of LSTM. \cite{yang2021spagan} sampled the shortest path based on their attention score to the central node and combined them in a mixed way. PPNP and APPNP\cite{klicpera2018predict} used the relationship between GCN and PageRank to derive an improved communication scheme based on personalized PageRank. 
	
	With the scale of graph data increasing, the full-batch algorithm is no longer applicable, and the mini-batch algorithm using stochastic gradient descent is applied. In recent years, some studies based on different sampling strategies have been proposed. \cite{hamilton2018inductive} tried random neighbor node sampling for the first time to limit large amounts of nodes caused by recursive node sampling. Layer-wise sampling techniques was applied in \cite{chen2018fastgcn,gao2018large,huang2018adaptive,ying2018hierarchical}, which only consider fixed number of neighbors in the graph to avoid ``neighbor explosion''. \cite{chen2018fastgcn,chen2018stochastic} used a control variate-based algorithm that can achieve good convergence by reducing the approximate variance. Besides, subgraph sampling was first introduced by \cite{zeng2019accurate}, which used a probability distribution based on degree. \cite{chiang2019cluster} performed clustering decomposition on the graph before the training phase and randomly sampled a subgraph as a training batch at every step. \cite{zeng2020graphsaint} sampled subgraph during training based on node and edge importance. All these sampling techniques are based on GCNs, how to extract deeper topology information from large graph datasets is still a problem.
	\section{Proposed method}
	In this section, we present the PathSAGE. Firstly, the sampling algorithm is introduced in 3.1. Secondly, we detail the aggregator in 3.2. Finally, we discuss the difference between our model and the related models.
	\begin{algorithm} 
		\SetAlgoLined
		\DontPrintSemicolon
		\caption{Random Path Sampling}  
			\KwIn{garph $G(V,E)$\
			central node $c$\\
			sample depth $s$\\
			sample num each length $L =\{n_1, n_2, …, n_s\}$\} }
			\KwOut{path sequcences with different length\\ 
			$\{\bm{P}_1, \bm{P}_2, ..., \bm{P}_s\}$}
			\SetKwFunction{Fun}{Ramdom Path Sampling}
			\Fun{$G$, $c$, $s$, $L$}{
				
			\ForEach{$l = 0 \to s$}{
			$i\gets0$;
			
			\While{$i < n_l$}{
			 $u\gets c$;
			 
			 $P\gets \{u\}$;
			 
			\For{$j = 0 \to l$}{
			 $u \gets$ Node randomly selected from $u$'s neighbors;
			 
			 $P\gets P \cup \{u\}$}
		 
			 $\bm{P}_l \gets \bm{P}_l \cup \{P\}$;
			 
			 $i \gets i+1$;}
		 
			}
		\Return $\{\bm{P}_1, \bm{P}_2, ..., \bm{P}_s\}$
		}
	\end{algorithm}
	\subsection{Path Sampling}
	Except deepening model, another way to expand receptive field of CNNs is to increase the size of convolution kernel. Following this idea, we sample the node sequences starting from central node and regard these paths as the context of the central node. Therefore, the receptive field can be expanded by extending these paths.
	
	We use a straightforward random sampling algorithm based on random walk, shown in \textbf{Algorithm 1}: for a central node, a sampling starts from it and randomly selects a neighbor node each time until reaching the preset path length, and multiple path sequences for the corresponding central node of various lengths can be obtained in this way, constituting a training sample.
	
	\begin{figure*}[htb]
		\begin{center}
			\includegraphics[height=4.5cm]{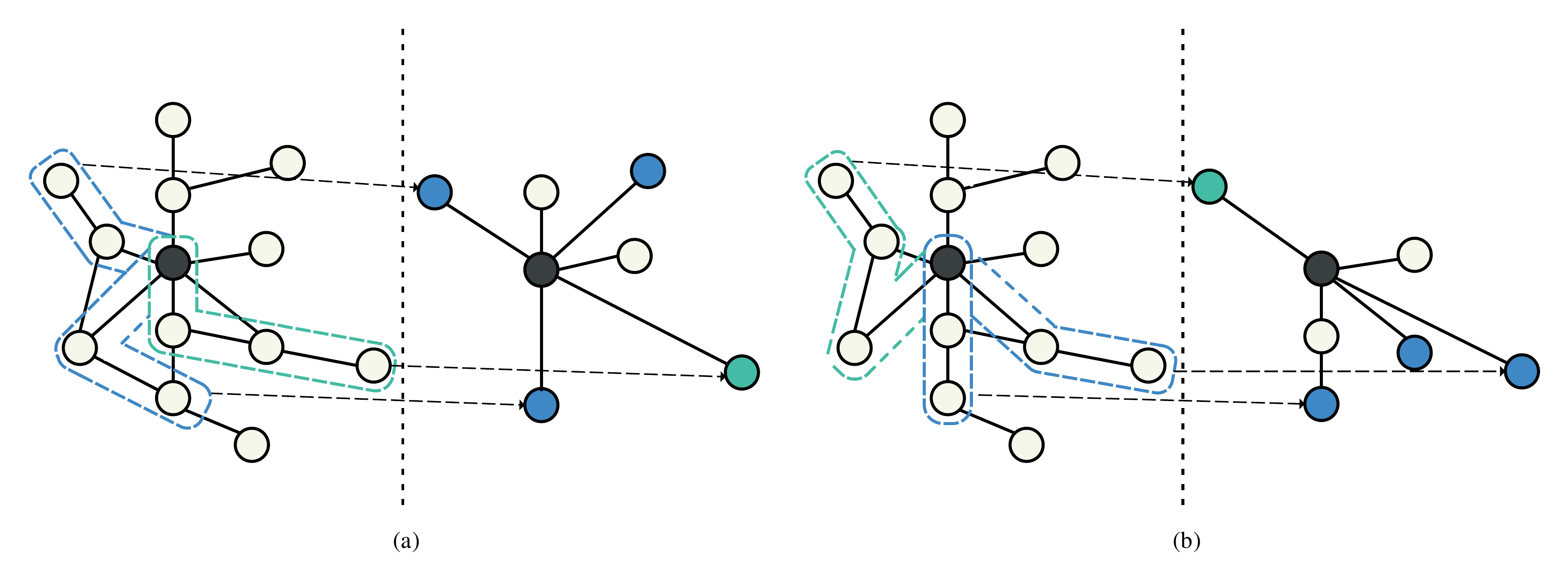}
		\end{center}
		\caption{Mechanism of sampling and aggregation. (a) and (b) are two different possible training samples with the same central node. Paths with same colors are with same lengths same length and share the same aggregators. }
		\label{fig1}
	\end{figure*}
	
	\begin{figure}[htbp]
		\begin{center}
			\includegraphics[height=4.5cm]{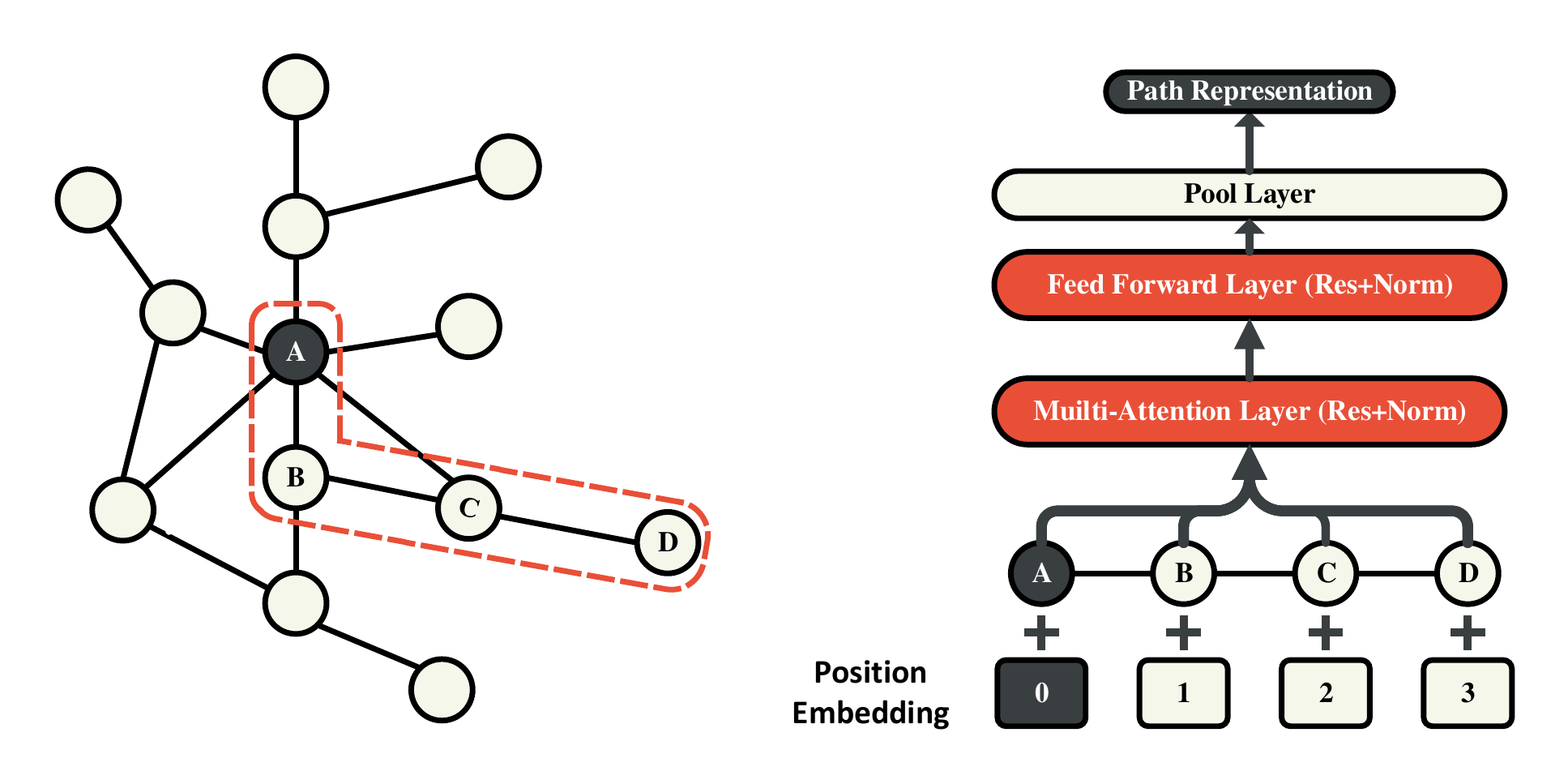}
		\end{center}
		\caption{Structure of first aggregation in aggregator. A specific sampled path’s features adds the position embedding, through Transformer encoder layers to fuse the information.
		}
		\label{fig2}
	\end{figure}
	
	\subsection{Aggregator}
	There are two aggregations in the aggregator: the first one aggregate path sequences; the second one aggregate different paths as the final representation of the central node. 	
	For the first aggregation, We formulate each path sequence as:
	\begin{equation}
		{\bm{P}}_{ij}^{}\; = \;\{ {\vec a_{ij}^1,\;\vec a_{ij}^2,\;\vec a_{ij}^3,\;...,\;\vec a_{ij}^j\;} \}
	\end{equation}
	
	To utilize the position information of path sequence, we define positions of nodes in paths as their distances to the central nodes. As same as Transformer, we add nodes' features and their positional vectors together as the input of the structure:
	\begin{equation}\label{eq6}
		\bm{\tilde{P}}_{ij}\; = \;\{ {a_{ij}^1\; + \;pos\_emb\left( 1 \right),...,\;a_{ij}^j\; + \;pos\_emb\left( j \right)} \}
	\end{equation}
	For the $pos\_emb(\cdot)$, we use $sine$ and $cosine$ functions of different frequencies to generate the positional vectors. In each dimension, the positional embedding is:
	
	\begin{equation}\label{eq7}
		pos\_emb(p)_{2i} = sin(\frac{{p}}{{{{10000}^{\frac{{2i}}{{{d}}}}}}})
	\end{equation}
	
	\begin{equation}\label{eq8}
		pos\_emb(p)_{2i + 1} = cos(\frac{{p}}{{{{10000}^{\frac{{2i}}{{d}}}}}})
	\end{equation}
	Where $p$ is the position in the sequences, $2i$ and $2i+1$ mean the odd and even dimensions of position embedding, and $d$ is the number of features. 
	
	After that, we apply the Transformer encoder on each path: 
	
	\begin{equation}\label{eq9}
		\bm{\tilde{P}}_{ij}^{k} = transformer\_bloc{k^k}(\bm{\tilde{P}}_{ij}^{k-1})
	\end{equation}
	Where $k$ means $k$-th Transformer encoder layer. Noted that, in a $m$-layer Transformer encoder, the output of the last layer $\bm{\tilde{P}}_{ij}^{m}$ is a sequence of features. We only take the output at position 0 as the final representation of the path sequence.
	
	\begin{equation}\label{eq10}
		\tilde{P}_{ij}' = \left[\bm{\tilde{P}}_{ij}^{m}\right]_{0}
	\end{equation}	
	
	Following these equations (\ref{eq6})-(\ref{eq10}), we can obtain representations of paths with different length to the central node, as shown in Fig. 1. In the second aggregation, we apply an average pooling layer to aggregate the central node paths. Then we concatenate all the path representation and apply a feed-forward layer with non-linearity to fuse these features. The final output of a central node $C'$ is computed as following:
	
	\begin{equation}
		\begin{array}{c}
			C = \;concat\left( {{C_1},\;...,\;{C_s}} \right)\\
			\\
			where\;{C_i}\; = \;Average\left( \tilde{P}_{i1}',\;...,\;\tilde{P}_{in}' \right)
		\end{array}
	\end{equation}
	
	\begin{equation}
		{C'}\; = \;max\left( {0,\;C{\bm{W}_1}\; + \,{b_1}} \right){\bm{W}_2}\; + \;{b_2}
	\end{equation}
	where $s$ denotes the sample depth, $n$ denote the number of paths sampled in 
	
	\subsection{Comparisons to related work}
	\begin{itemize}
		\item We introduce a sequence transduction model into our structure, but it is distinct from related work based on these models. LSTM\cite{hochreiter1997long} was also used in GraphSAGE\cite{zeng2020graphsaint} to aggregate node’s features, it is very sensitive to the order of the input sequence. In contrast, neighbor nodes are disordered, and the authors have rectified it by consistently feeding randomly-ordered sequences to the LSTM. In our model, paths to the central node already have orders, which are naturally suitable for sequence processing model. 
		\item The difference between our attention mechanism and GAT is that we collect all the mutual attention information of each node in the path. The output of our model is integrated information of entire sequence, instead of only attend to the central node.
		\item Path sampling method is also used in SPAGAN\cite{yang2021spagan}. But each sampled path in SPAGAN is the shortest one to the central node, this sample technique will leads to a high computational overhead and limits the ability of the model to be applied to large graph datasets. By contrast, we sample paths randomly, which greatly save the computation overhead. Besides, in our sample algorithm, the same node may have a different path to the central node, which may help the model acquire more diverse patterns around the central node while saving the computational overhead. 
	\end{itemize}
	\section{Experiments}
	In this section, we introduce the datasets and the experiment setting in 4.1 and 4.2 respectively. we present the results of evaluation in 4.3. 
	\begin{table}[htbp]
		\caption{Summary of inductive learning tasks' statistics.}
		\begin{center}
			\centering
			\renewcommand{\arraystretch}{2.2}
			\setlength{\tabcolsep}{4mm}{
				\begin{tabular}{|c|c|c|c|}
					\hline
					&     \textbf{Reddit} &       \textbf{Yelp} &     \textbf{Flickr} \\
					\hline
					type & single-label & multi-label & multi-label \\
					
					\#Node &    232,965 &    716,847 &     89,250 \\
					
					\#Edges & 11,606,919 &  6,977,410 &    899,756 \\
					
					\#Features &        602 &        300 &        500 \\
					
					\#Classes &         41 &        100 &          7 \\
					
					\makecell[c]{Train / Val \\/ Test} & \makecell[c]{66\% / 10\% \\/ 24\%} & 	\makecell[c]{75\% / 10\% \\/ 15\%} & \makecell[c]{50\% / 25\% \\/ 25\%} \\
					\hline
			\end{tabular} }
		\end{center}
		\label{tab2}
	\end{table}
	\subsection{Dataset}
	Three large-scale datasets are used to evaluate our model: 1) \textbf{Reddit}\cite{hamilton2018inductive}: a collection of monthly user interaction networks from the year 2014 for 2046 sub-reddit communities from Reddit that connect users when one has replied to the other, 2) \textbf{Flickr}: a social network built by forming links between images sharing common metadata from Flickr. 3) \textbf{Yelp}\cite{zeng2020graphsaint}: a network that links up businesses with the interaction of its customers. Reddit is a multiclass node-wise classification task; Flickr and Yelp are multilabel classification tasks. 
	The detail statistics of these datasets are shown in Tab 1.
	
	\begin{table}[t]
		
		\caption{Perforamnce on inductive learning tasks(Micro-F1).}
		\begin{center}
			\centering
			
			\renewcommand{\arraystretch}{2}
			\setlength{\tabcolsep}{4mm}{
				\begin{tabular}{|c|c|c|c|}
					\hline
					\textbf{Model} &     \textbf{Reddit} &       \textbf{Yelp} &     \textbf{Flickr} \\
					\hline
					GCN & 0.933$\pm$0.000 & 0.378$\pm$0.001 & 0.492$\pm$0.003 \\
					
					GraphSAGE & 0.953$\pm$0.001 & 0.634$\pm$0.006 & 0.501$\pm$0.013 \\
					
					FastGCN & 0.924$\pm$0.001 & 0.265$\pm$0.053 & 0.504$\pm$0.001 \\
					
					S-GCN & 0.964$\pm$0.001 & 0.640$\pm$0.002 & 0.482$\pm$0.003 \\
					
					AS-GCN & 0.958$\pm$0.001 &     -      & 0.504$\pm$0.002 \\
					
					ClusterGCN & 0.954$\pm$0.001 & 0.609$\pm$0.005 & 0.481$\pm$0.005 \\
					
					GraphSAINT & 0.966$\pm$0.001 & \bf{0.653$\pm$0.003} & 0.511$\pm$0.001 \\
					
					ours & \bf{0.969$\pm$0.002} & 0.642$\pm$0.005 & \bf{0.511$\pm$0.003} \\
					\hline
			\end{tabular} }
			\label{tab4}
		\end{center}
	\end{table}
	\subsection{Experiment setup}
	
	We build our model on Pytorch framework\cite{paszke2017automatic} and construct the Transformer encoder based on UER\cite{zhao2019uer}. For all tasks, We train the model with Adam SGD optimizer\cite{kingma2014adam} with learning rate 1e-3 and a learning rate scheduler with 0.1 warmup ratio. We use two layers Transformer, each of which has 8 attention heads to gather the features of paths. The batch size is 32. The dropout layers are applied between each sub-layer in Transformer layer. We use different dropout rates in the output layer and Transformer encoder, which are 0.3 and 0.1 respectively. We set the sampling length of the path ranging from 1 to 8 (depth $s = 8$), and the number of paths sampled in each length are [5, 5, 5, 5, 5, 10, 10, 10]. For multi-label classification task Flickr and Yelp, the final output is obtained through a $sigmoid$ activation, and in Reddit, the final output is obtained through a $softmax$ activation. The hidden dimension is 128 for Reddit and Flickr, 512 for Yelp. 
	
	\subsection{Result}
	We compare our model with seven state-of-the-art model:  GCN\cite{kipf2017semisupervised}, GraphSAGE\cite{hamilton2018inductive}, FastGCN\cite{chen2018fastgcn}, S-GCN\cite{chen2018stochastic}, AS-GCN\cite{huang2018adaptive}, ClusterGCN\cite{chiang2019cluster}, GraphSAINT\cite{zeng2020graphsaint}. GraphSAGE uses a random node sampling and LSTM aggregator. FastGCN and S-GCN use a control variate-based algorithm that can achieve good convergence by reducing the sampling variance. ClusterGCN performs clustering decomposition on the graph before the training phase and randomly sampling a subgraph as a training-batch at every step. GraphSAINT samples subgraph while training with several sampling methods based on node and edge importance. The evaluation on these tasks uses the $Micro-F1$ metric, and report the mean and confidence interval of the metrics by five runs.
	
	The results of inductive learning experiments are shown in Tab 2. As we can see from the table, For Reddit, our model outperforms all the baseline models. For Flickr, we achieve a comparable F1-score with the top-performing model. For Yelp, we surpass most GCN models, second only to GraphSAINT. One hypothesis to explain the difference of the results is: Reddit and Flickr have more training samples and number of nodes features, which makes the attention mechanism have enough data to capture the relationships between nodes.
	
	\begin{figure*}[t]
		\begin{center}
			\includegraphics[height=4.7cm]{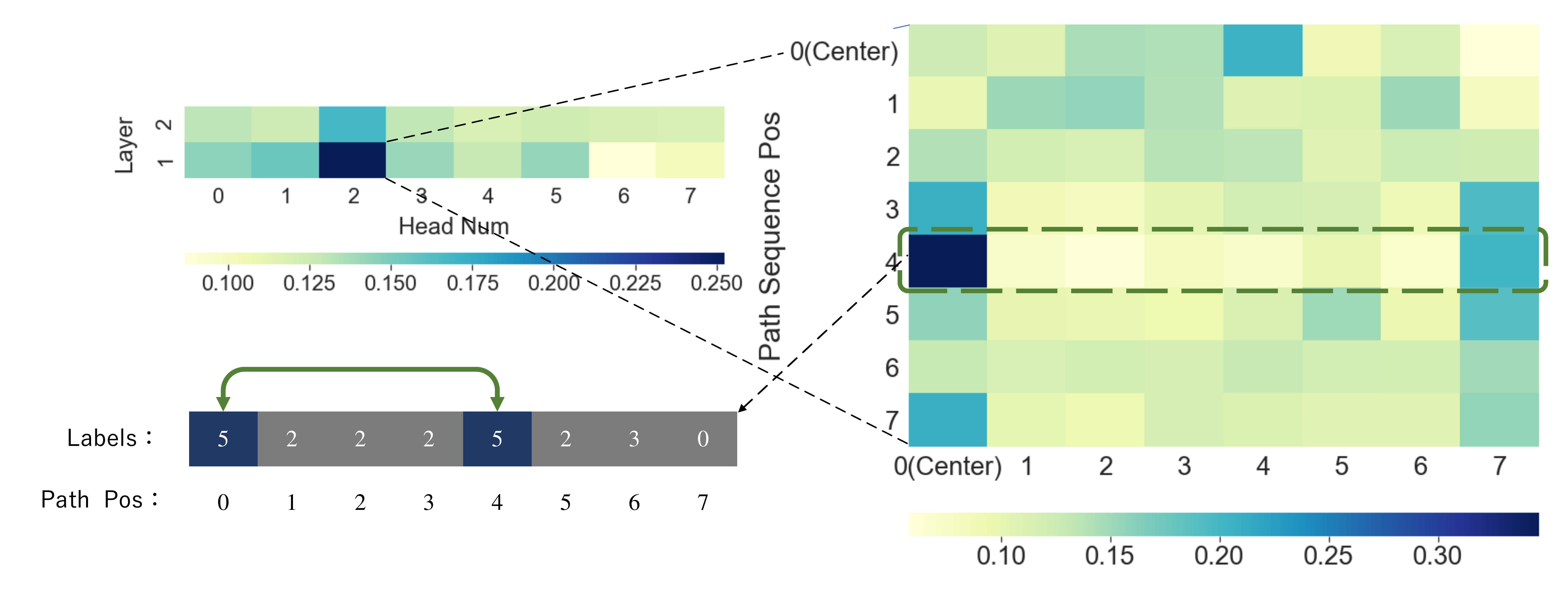}
		\end{center}
		\caption{Detection of attention weights in Transformer that node have the same label in the path sequence may receive higher weights at some attention heads. The heatmap in the upper-left corner is the weight of position 4 of different layers attend to position 2 and attention heads, the right one is the complete heatmap of the selected head. The lower-left picture shows the labels of the nodes in the sequence. }
		\label{fig3}
	\end{figure*}
	
	\section{Attention Analysis}
	We observed the attention weight of the trained model in the PubMed test set. At some attention heads, we find that nodes with the same labels get very high attention scores to each other. We visualize an example in Fig 3. Central node 0 and node 4 have the same label and receive extremely high attention scores on second attention heads of the first layer. By observing the attention scores for the whole sequence, we can see that this score also occupies a significant share in the sequence. 
	
	This observation proves that the attention mechanism can successfully capture the information that is helpful to the downstream tasks in the aggregation of paths.
	
	\section{Conclusion}
	In this paper, we propose a model---PathSAGE, which can expand the size of receptive field without stacking layers which solved the problem of ``neighbor explosion'' and ``over smoothing''. Regarding all paths as the neighbor nodes of the central node, PathSAGE samples paths starting from the central node based on random walk and aggregates these features by a strong encoder---Transformer. Consequently, the model can obtain more features of neighbor nodes and more patterns around the central node. In our experiment, PathSAGE achieves the state-of-the-art on inductive learning tasks. Our model provides a novel idea to handle large-scale graph data.
	
	\section*{Acknowledgments}.
	This work is supported by National Key R\&D Program of China (No.2017YFB1402405-5), and the Fundamental Research Funds for the Central Universities (No.2020CDCGJSJ0042). The authors thank all anonymous reviewers for their constructive comments.


\begin{thebibliography}{10}
		\providecommand{\url}[1]{\texttt{#1}}
		\providecommand{\urlprefix}{URL }
		\providecommand{\doi}[1]{https://doi.org/#1}
		
		\bibitem{bruna2014spectral}
		Bruna, J., Zaremba, W., Szlam, A., LeCun, Y.: Spectral networks and locally
		connected networks on graphs (2014)
		
		\bibitem{chen2018stochastic}
		Chen, J., Zhu, J., Song, L.: Stochastic training of graph convolutional
		networks with variance reduction (2018)
		
		\bibitem{chen2018fastgcn}
		Chen, J., Ma, T., Xiao, C.: Fastgcn: Fast learning with graph convolutional
		networks via importance sampling (2018)
		
		\bibitem{chiang2019cluster}
		Chiang, W.L., Liu, X., Si, S., Li, Y., Bengio, S., Hsieh, C.J.: Cluster-gcn: An
		efficient algorithm for training deep and large graph convolutional networks.
		In: Proceedings of the 25th ACM SIGKDD International Conference on Knowledge
		Discovery \& Data Mining. pp. 257--266 (2019)
		
		\bibitem{defferrard2017convolutional}
		Defferrard, M., Bresson, X., Vandergheynst, P.: Convolutional neural networks
		on graphs with fast localized spectral filtering (2017)
		
		\bibitem{gao2018large}
		Gao, H., Wang, Z., Ji, S.: Large-scale learnable graph convolutional networks.
		In: Proceedings of the 24th ACM SIGKDD International Conference on Knowledge
		Discovery \& Data Mining. pp. 1416--1424 (2018)
		
		\bibitem{hamilton2018inductive}
		Hamilton, W.L., Ying, R., Leskovec, J.: Inductive representation learning on
		large graphs (2018)
		
		\bibitem{hochreiter1997long}
		Hochreiter, S., Schmidhuber, J.: Long short-term memory. Neural computation
		\textbf{9}(8),  1735--1780 (1997)
		
		\bibitem{huang2018adaptive}
		Huang, W., Zhang, T., Rong, Y., Huang, J.: Adaptive sampling towards fast graph
		representation learning. arXiv preprint arXiv:1809.05343  (2018)
		
		\bibitem{kingma2014adam}
		Kingma, D.P., Ba, J.: Adam: A method for stochastic optimization. arXiv
		preprint arXiv:1412.6980  (2014)
		
		\bibitem{kipf2017semisupervised}
		Kipf, T.N., Welling, M.: Semi-supervised classification with graph
		convolutional networks (2017)
		
		\bibitem{klicpera2018predict}
		Klicpera, J., Bojchevski, A., G{\"u}nnemann, S.: Predict then propagate: Graph
		neural networks meet personalized pagerank. arXiv preprint arXiv:1810.05997
		(2018)
		
		\bibitem{li2018deeper}
		Li, Q., Han, Z., Wu, X.M.: Deeper insights into graph convolutional networks
		for semi-supervised learning. In: Proceedings of the AAAI Conference on
		Artificial Intelligence. vol.~32 (2018)
		
		\bibitem{liu2019geniepath}
		Liu, Z., Chen, C., Li, L., Zhou, J., Li, X., Song, L., Qi, Y.: Geniepath: Graph
		neural networks with adaptive receptive paths. In: Proceedings of the AAAI
		Conference on Artificial Intelligence. vol.~33, pp. 4424--4431 (2019)
		
		\bibitem{monti2017geometric}
		Monti, F., Boscaini, D., Masci, J., Rodola, E., Svoboda, J., Bronstein, M.M.:
		Geometric deep learning on graphs and manifolds using mixture model cnns. In:
		Proceedings of the IEEE conference on computer vision and pattern
		recognition. pp. 5115--5124 (2017)
		
		\bibitem{paszke2017automatic}
		Paszke, A., Gross, S., Chintala, S., Chanan, G., Yang, E., DeVito, Z., Lin, Z.,
		Desmaison, A., Antiga, L., Lerer, A.: Automatic differentiation in pytorch
		(2017)
		
		\bibitem{vaswani2017attention}
		Vaswani, A., Shazeer, N., Parmar, N., Uszkoreit, J., Jones, L., Gomez, A.N.,
		Kaiser, L., Polosukhin, I.: Attention is all you need (2017)
		
		\bibitem{velivckovic2017graph}
		Veli{\v{c}}kovi{\'c}, P., Cucurull, G., Casanova, A., Romero, A., Lio, P.,
		Bengio, Y.: Graph attention networks. arXiv preprint arXiv:1710.10903  (2017)
		
		\bibitem{yang2021spagan}
		Yang, Y., Wang, X., Song, M., Yuan, J., Tao, D.: Spagan: Shortest path graph
		attention network (2021)
		
		\bibitem{ying2018hierarchical}
		Ying, R., You, J., Morris, C., Ren, X., Hamilton, W.L., Leskovec, J.:
		Hierarchical graph representation learning with differentiable pooling. arXiv
		preprint arXiv:1806.08804  (2018)
		
		\bibitem{zeng2019accurate}
		Zeng, H., Zhou, H., Srivastava, A., Kannan, R., Prasanna, V.: Accurate,
		efficient and scalable graph embedding. In: 2019 IEEE International Parallel
		and Distributed Processing Symposium (IPDPS). pp. 462--471. IEEE (2019)
		
		\bibitem{zeng2020graphsaint}
		Zeng, H., Zhou, H., Srivastava, A., Kannan, R., Prasanna, V.: Graphsaint: Graph
		sampling based inductive learning method (2020)
		
		\bibitem{zhang2018gaan}
		Zhang, J., Shi, X., Xie, J., Ma, H., King, I., Yeung, D.Y.: Gaan: Gated
		attention networks for learning on large and spatiotemporal graphs (2018)
		
		\bibitem{zhao2019uer}
		Zhao, Z., Chen, H., Zhang, J., Zhao, X., Liu, T., Lu, W., Chen, X., Deng, H.,
		Ju, Q., Du, X.: Uer: An open-source toolkit for pre-training models.
		EMNLP-IJCNLP 2019 p.~241 (2019)
		
		\bibitem{zhou2018graph}
		Zhou, J., Cui, G., Zhang, Z., Yang, C., Liu, Z., Wang, L., Li, C., Sun, M.:
		Graph neural networks: A review of methods and applications. arXiv preprint
		arXiv:1812.08434  (2018)
		
	\end{thebibliography}
\end{document}